\pdfoutput=1

\documentclass[letterpaper, 10 pt, conference]{ieeeconf} \IEEEoverridecommandlockouts
\let\labelindent\relax

\usepackage{cite}
\usepackage{amsmath,amssymb,amsfonts}
\usepackage{graphicx}
\usepackage{textcomp}
\usepackage{enumitem}
\usepackage{xcolor}
\usepackage{tikz}
\usepackage{verbatim}
\usetikzlibrary{shapes.gates.logic.US,trees,positioning,arrows}
\usepackage{todonotes}
\usepackage{algorithmicx}
\usepackage[ruled]{algorithm}
\usepackage{algpseudocode}

\usepackage{multirow}

\usepackage{caption}
\usepackage{subcaption}
\DeclareCaptionLabelSeparator{periodspace}{.\quad}
\graphicspath{{./images/}}

\def\BibTeX{{\rm B\kern-.05em{\sc i\kern-.025em b}\kern-.08em
    T\kern-.1667em\lower.7ex\hbox{E}\kern-.125emX}}
\begin{document}

\definecolor{grey}{RGB}{190, 190, 190} 
\definecolor{tempexp}{HTML}{9262d1}
\definecolor{tempnot}{HTML}{fde910}
\definecolor{running}{HTML}{3283a8}

 \newcommand{\lorenzo}[1]{\textcolor{red}{#1}}

\title{\huge Task Planning with Belief Behavior Trees
\thanks{
$^1$The authors are with Istituto Italiano di Tecnologia, Genoa, Italy. \newline
$^2$Evgenii Safronov is also with Department of Informatics, Bioengineering, Robotics and Systems Engineering, Universit\`a di Genova, Genova, Italy
\texttt{evgenii.safronov@iit.it}
}
}

\author{{Evgenii Safronov$^{1,2}$, Michele Colledanchise$^1$, and Lorenzo Natale$^1$}}

%

\newcommand{\mnt}{\emptyset}
\newcommand{\msR}{\mathbb{R}}
\newcommand{\msS}{\mathbb{S}}
\newcommand{\msF}{\mathbb{F}}

\newcommand{\nt}{$\emptyset$}
\newcommand{\sR}{$\mathbb{R}$}
\newcommand{\sS}{$\mathbb{S}$}
\newcommand{\sF}{$\mathbb{F}$}

\maketitle

\begin{abstract}
In this paper, we propose \emph{Belief Behavior Trees} (BBTs), an extension to Behavior Trees (BTs) that allows to automatically create a  policy that controls a robot in partially observable environments. We extend the semantic of BTs to account for the uncertainty that affects both the conditions and action nodes of the BT.
The tree gets synthesized following a planning strategy for BTs proposed recently: from a set of goal conditions we iteratively select a goal and find the action, or in general the subtree, that satisfies it. Such action may have preconditions that do not hold. For those preconditions, we find an action or subtree in the same fashion. We extend this approach by including, in the planner, actions that have the purpose to reduce the uncertainty that affects the value of a condition node in the BT (for example, turning on the lights to have better lighting conditions). We demonstrate that \emph{BBTs} allows task planning with  non-deterministic outcomes for actions.  We provide experimental validation of our approach in a real robotic scenario and -- for sake of reproducibility -- in a simulated one.
\end{abstract}

\section{Introduction}
\label{sec:intro}
The video game industry proposed Behavior Trees (BTs) as an alternative to statecharts to describe the behavior of non-playable characters. The industry found BTs  successful to the point that they became an established tool appearing in textbooks \cite{millington2009artificial,rabin2014gameAiPro, BTBook} and game-coding software such as Pygame, Craft AI, and Unreal Engine. In robotics, both academia and industry follow a similar trend and use BTs to describe complex behaviors in a compact way. Moreover, BTs generalize other successful control architectures such as statecharts, the Subsumption architecture \cite{brooks1986robust}, and the Teleo-reactive Paradigm~\cite{BTBook}. 

\begin{figure}[t!]
    \centering
    \begin{subfigure}[t]{0.485\columnwidth}
        \centering
\includegraphics[width=0.8\columnwidth]{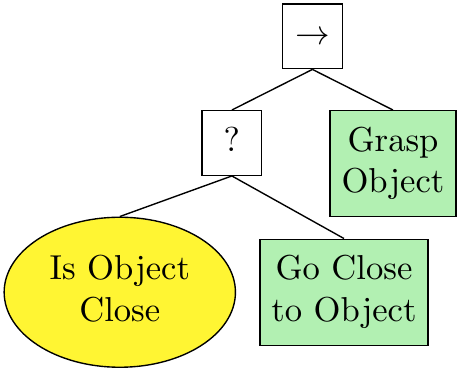}
         \caption{BT with classic semantic of condition nodes.}
         \label{IN.fig.front.classic}
    \end{subfigure}%
    ~
    \begin{subfigure}[t]{0.485\columnwidth}
\includegraphics[width=\columnwidth]{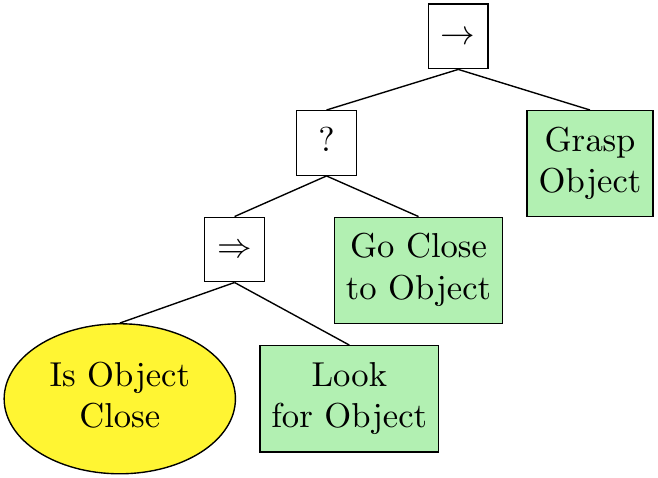}
   \caption{BT with new semantic of condition nodes.}
            \label{IN.fig.front.new}
    \end{subfigure}
    \vspace{0em}
    \caption{
    Examples of BTs using the classic (left) and the new (right) semantic for conditions. The syntax and semantic are described later in this paper.  }
    \label{IN.fig.front}
\end{figure}

The particular syntax and semantic of BTs, which will be described later, allow a BT to continuously check a set of conditions to evaluate the actions to be performed and the ones to be aborted, if any. However, in the classical semantic of BT, the designer assumes that the values of such conditions are either true or false. While this appears like a natural way to describe policies, it underlies an assumption: the conditions involve observable variables. Consider the example in Figure~\ref{IN.fig.front.classic}, the BT encodes the behavior that can be verbally described: as \emph{whenever the robot is close to the object, grasp it, when the robot is not close to the object, go close to it}. Such behavior makes sense as long as the robot can observe the object position, so that the BT can evaluate the condition \emph{object close}.

The real world remains intrinsically non-observable. However, in some cases, some variables can be observed, after performing specific actions or, in general, on a specific sub-space in the state space. For example, the condition \emph{object close} becomes observable only after the robot finds the object or if the robot knows the object position a priori.

In this paper, following the recent advances of BT semantic \cite{abtm}, we allow the conditions to be either true, false, or \emph{unknown}. Doing so, we can construct a new BT (in Figure~\ref{IN.fig.front.new}) that, in addition to the functionalities of the original BT, \emph{it performs a action that looks for the object whenever the value of its position relative to the robot is unknown}. Inspired by our recent work \cite{colledanchise2019towards}, a task planner generates such BT. The resulting BT performs both \emph{actuation}, which change the state space to achieve the goal and \emph{perception}, which makes observations on the robot's space. For the variables that are latent, the task planner operates directly on the \emph{belief state}, a state-space of the probability distribution over physical states.

Given the initial physical state of the system (with possibly unobserved conditions) and a goal definition, the algorithm
refines the tree until it results in successful execution.
Having moved from physical to \emph{belief} states, on each planning step we are looking for the \emph{most probable} unsatisfied condition.
If the condition holds \emph{false}, we insert an \emph{actuation} action, if it is \emph{unobserved}, we insert a \emph{perception} action.
As both actuations and perceptions have probabilistic outcomes, we developed an extention of standard BTs, which we called \emph{Belief Behavior Trees} (BBTs).
BBTs allows to perform planning in the belief space, by applying mulitple ticks, therefore simulating all possible scenarios of execution and computing the \emph{probability of success} of each execution trace.

%
%

To summarize, in this paper we combine recent advances in the BT community to take a  step towards BT planning in partially-observable environments.
We show how to handle uncertainty in the BT formulation, then propose the \emph{BBT}, a BT that allows planning with the non-deterministic outcomes for actions and conditions. We provide experimental validation running our approach in a real robotic scenario and on a simulation.  For reproducibility, we also make available online the source code of our framework and the simulated scenario.

\section{Related Works}
\label{sec:related_work}

The solution to long-horizon task planning in uncertain environments lies beyond the state of the art~\cite{kaelbling2013integrated, ghallab2016automated}. Early works apply the \emph{determinize-and-plan} approach where they compute the most likely physical state and then plan in a deterministic domain from that state. However, such approaches have a fundamental issue, they cannot consider actions that reduce the uncertainty as the planner runs on a deterministic domain. Later works include perception actions in the planner, however, they assume that the future observations are always the most probable ones~\cite{platt2010belief}. Other works\cite{hadfield2015modular}, extend the classical planning operators~\cite{ghallab2016automated} defining preconditions and effects in the belief space to use off-the-shelf task planners to operate in belief space. However, they still rely on the maximum likelihood observation assumption above.
Kaelbling et al.~\cite{kaelbling2013integrated} outlined a framework that blends acting and planning to handle uncertainty in robot tasks. They construct task plans in belief space under maximum likelihood observation assumption.

Regarding the automatic creation of BTs, the research community follows two main directions: a learning one, where they construct the BTs in a model-free fashion optimizing over a reward function, and a task planning one, where they construct the BTs in a model-based fashion.
Works include the use of Reinforcement Learning techniques~\cite{banerjee2018autonomous} and in particular genetic programming to build and combine BTs selecting the ones that have the highest reward~\cite{jones2018evolving}. Other works~\cite{colledanchise2015learning} construct the BT by mixing a greedy approach with genetic programming to maximize the reward function. Recent works combine the manual design of BTs with machine learning techniques to learn policies with the desired performance~\cite{sprague2018adding}. Other works synthesize BTs from demonstration~\cite{paxton2017costar,french2019learning}.
The employment of task planning approaches is a recent trend, early works define a systematic framework to automatically generate complex BT structures from a repository of simpler ones. However, that approach strongly depends on a repository of hand-made structures (in a similar fashion of Hierarchical Task Network), whereas our approach automatically creates BTs from a set of simple actions and their pre- and postconditions.
Other works define the goal in the form of a Linear Temporal Logic specification and then employ a verification tool to synthesize a BT that is correct by construction~\cite{colledanchise2017synthesis}. Later works propose the construction of the BT  based on the idea
of backchaining. Starting from the goal condition the algorithm finds actions that meet those conditions~\cite{colledanchise2019towards}.

The fundamental difference with all the work above on automatic synthesis of BTs lies in the fact that we do consider uncertainty in both the actions outcome (i.e. actuation and perception) and the value of condition nodes.

\section{Background}
In this section, we briefly describe the syntax and semantic of BTs. We follow conventional BT syntax with few important differences. The literature includes detailed descriptions~\cite{BTBook,abtm}.

We use the common definitions of  \emph{parent} and \emph{child} for the tree structure. The root is the node without parents, the nodes have only one parent. Graphically, the children of nodes are placed below it, as shown in Figure~\ref{IN.fig.front}. The children are executed in the order from left to right.

The execution of a BT begins from the root node. It sends \emph{ticks}~\footnote{A tick is a signal that allows the execution of a child.} its children, from left to right. When a parent sends a tick to a child, the child can be executed. The child returns to the parent a status \emph{running} (\sR) if its execution has not finished yet, \emph{success} (\sS) if it has achieved its goal, or \emph{failure} (\sF) otherwise.

We describe the semantic of the nodes we use in this paper.

\paragraph*{Fallback}
The fallback is a control node that returns status success \sS~(running \sR) as soon as it finds a child that returns success \sS~(running \sR). It returns failure \sF~only if all the children return failure \sF. When a child returns running or success, the fallback node does not tick the next child (if any).
The fallback node is represented by a box with a ``?", as in Figure~\ref{IN.fig.front}.

\paragraph*{Sequence}
The sequence node returns failure (running) as soon as it finds a child that returns failure (running). It returns success only if all the children return success. When a child returns running or failure, the sequence node does not tick the next child (if any). The sequence node is represented by a box with a ``$\rightarrow$", as in Figure~\ref{IN.fig.front}.

\paragraph*{Skipper}
The skipper node return success (failure) if a child return success (failure) otherwise it ticks the next child. The skipper node is represented by a box with a ``$\Rightarrow$", as in Figure~\ref{IN.fig.front}.

\paragraph*{Action}
The action node returns success if the action is completed and failure if the action cannot be completed. Otherwise, it returns
 running.  An action node is represented as a rectangle, as in Figure~\ref{IN.fig.front}.
\paragraph*{Condition}
The condition node checks if a condition is satisfied or not, returning success or failure accordingly. In this paper we allow conditions to return running to encode a condition whose value is unknown, as described in Section~\ref{sec:intro}. A condition node  {is represented as an ellipse, as in Figure~\ref{IN.fig.front}.

\section{Problem Formulation}
Conditions are defined by their literal, e.g., $luminousity\_ok$ may hold a value from the set of $\{true$, $false$, $unknown\}$.
Actions are described by their set of preconditions and action outcomes. The set of preconditions $C_{pre}^a$ for the action $a$ is a set of (condition, value) pairs. 
An action could be only executed if all its preconditions hold.
Action outcomes are probabilistic distribution over postcondition sets, i.e. for action $a$ we can have\footnote{Here and after $\bigcup_i p_i A_i$ denotes a discrete probability distribution of $A$ with probabilities $p_i$.}  $\bigcup_i p_i C_{post_i}^a$.
The action execution results in a change of some condition variables. Hence the problem can be stated as follows: given a set of actions defined above and a set of goal conditions, define a task planning algorithm that builds a BT policy that satisfies the goal conditions with no less than target probability.

\section{Handling Uncertainty in Behavior Trees}
In this section, we present the first contribution of this paper.
We show how to handle two types of uncertainty in the BT formulation: the \emph{current state uncertainty} and the \emph{future state uncertainty}.
\subsection{Handling current state uncertainty in BTs}
In many real-world scenarios, the robot only partially observes the environment.
Geometrical occlusions (e.g., doors, furniture, other objects), poor signal-to-noise ratio (e.g., low luminosity for the vision and high noise level for sound recording), and the sensor noise cause the current state uncertainty.

From a task planning standpoint, a set of conditions describes the current state.
Such conditions take values either \emph{true} or \emph{false}.
These conditions correspond to condition nodes of BT, which return respectively \emph{success} or \emph{failure}.
However, in the case of partially-observable environments, the value of a condition can be unknown.
Looking back at the example in Figure~\ref{IN.fig.front}, if we allow conditions in the BT to return \emph{running} status whenever the value of the condition is unknown, then we can use the Skipper node~\cite{abtm} to execute the observation. Such an observation could be implemented in form of a single action or as a subtree of BT.

%
%

\subsection{Handling future state uncertainty in BTs}

An action could result in different possible outcomes. For example, an attempt to grasp a bottle from a table could either result in a successful grasp, in a minor failure (e.g. the bottle stays on a table state of the world is unchanged), or in a severe failure (e.g. the bottle falls on the floor).

In this paper, we describe actions along with the \textit{preconditions} (i.e., the set of conditions that must be true before their execution) and the \textit{postconditions} (i.e., the set of conditions that must be true after the action terminates).

To account for future state uncertainty, we define a set of all the possible postconditions.
    Assuming that the possible postconditions are independent events, we can add a probability of each postconditions set and define action outcome as a probabilistic sum:
    \begin{multline}
    \hat{a} := \{C^a_{pre} , C^a_{post} \} \longrightarrow \hat{a} := \{ C^a_{pre}, \bigcup_i p^a_i {C^a_{post} }_i \} \\
    C^a_{pre} \subset C, C^a_{post} \subset C , {C^a_{post} }_i \subset C
        \label{eq:actions-postc}
    \end{multline}
    where $C$ is a set of all conditions, $C^a_{pre}$, is the set of precondition, and $C^a_{post}$ is the set of postconditions. In this work, we do not discuss the origin of these probabilities. They can be learned automatically during the task execution or calculated from robot safety-related and performance parameters. 
%
    \subsection{Uncertainty entanglement}
    It is important to highlight that current and future state uncertainties are \emph{not} independent in the general case.
    We need to point out cases when current state uncertainty leads to future state uncertainty and vice versa.
    \paragraph*{Perception actions}
    Sometimes a robot can make an extra action to resolve a current state uncertainty.
    For example, it might open a fridge to check if $in\_fridge(soda)$  is true.
    It makes sense to perform this action only in case if the value of condition above is $unknown$. 
    Hence, observations transform current state uncertainty into future state uncertainty.

    \begin{table}[h]
        \centering
    \begin{tabular}{|l | c | c|}
    \hline
                    	& Preconditions & Postconditions 	        \\ \hline
    look for object         &  $in\_fridge(obj)	= \msR$    &  $0.5, in\_fridge(obj)	= \msS$ \\
    in fridge     	              &  	            &  $0.5, in\_fridge(obj)	= \msF$       \\ \hline
    grasp 	         		& ...       	& $1, grasped(obj) = \msR$\\ \hline
    after\_grasp\_check           	&  $grasped(obj) = \msR$ & $0.8, grasped(obj) = \msF$\\
                       &        & $0.2, grasped(obj) = \msS$   \\ \hline
    \end{tabular}
        \caption{Uncertainty entanglement examples.}
        \label{tbl:uncertainty-actions}
    \end{table}
    \paragraph*{Actions with unknown outcomes}

    Another important example is the action \emph{grasp}.
    For some robots, it is possible to immediately sense if the object was successfully grasped (e.g., by tactile sensors).
    For others, the result of grasping is \emph{unknown} until we make an \emph{observation} (e.g., through putting the object close to the robot's camera for correct recognition).
    In this case, the action \emph{grasp} can have only one postcondition, but with \emph{unknown} value\footnote{Recall that logic states true, false or unknown are mapped to success (S), failure (F) and running (R) respectively in the BT} (see Table \ref{tbl:uncertainty-actions}).
    \par
    Note, that if our grasping performs well, the success probability of $after\_grasp\_check$ observation should be significantly higher than the one for an arbitrary observation.
    To fix that, one can use additional postcondition $after\_grasp = \msS$ for $grasp$ and use it as a precondition for $grasp\_check$ to enforce this type of observation.
    Another way to solve it is to put both actions (actuation and sensing) together in a sequence making a sub-BT or node template, to let the robot always make an observation after grasping.

    \section{Belief Behavior Tree}
    In this section, we present the second contribution of this paper. We define  BBTs, an extension to the definition of BT where conditions return running whenever the value of the condition is not known and actions have nondeterministic execution outcome.
    The planning will result in the construction of BBT.


    \paragraph*{Belief state in BTs}
    A physical state contains all conditions and their values.
    Belief state $m$ is defined as a probabilistic distribution of physical states $s_i$ as follows.
    $$
    m := \bigcup_i{p_i s_i}
    $$
where $p_i$ is the probability of being in state $s_i$.
    \begin{figure*}[ht]
        \centering
        \includegraphics[width=0.95\textwidth]{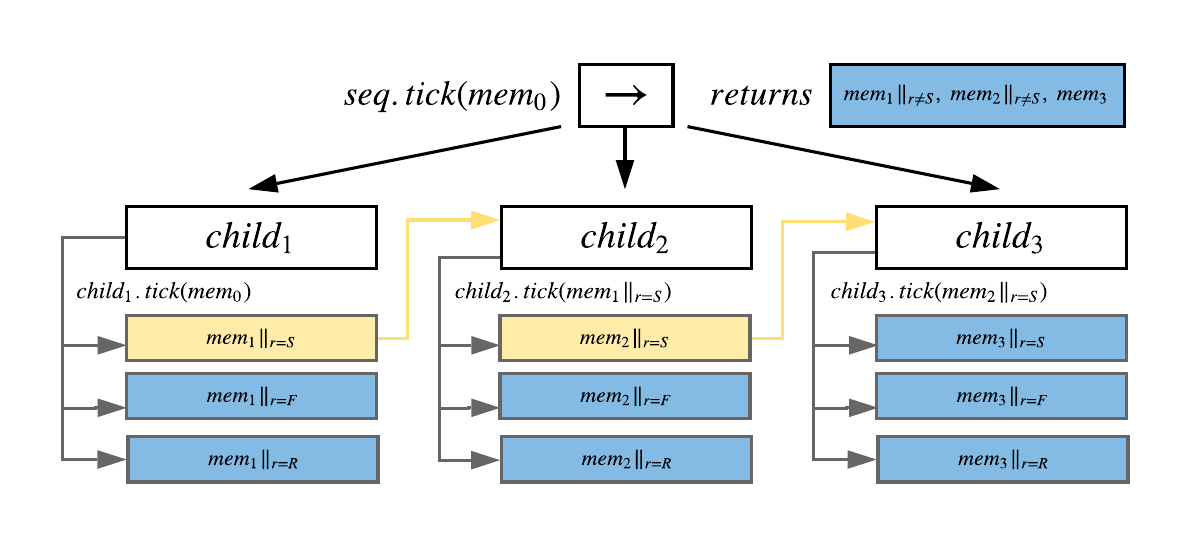}
        \caption{A belief state flow chart example for a tick execution in a Sequence control node. Physical states highlighted in blue forms the returned belief state by a sequence.}
        \label{fig:seq_tick}
    \end{figure*}
    \subsection{Leaf nodes}
    We now describe how actions and conditions and their corresponding BBT nodes act on a belief state.
    \paragraph*{Actions}
    Deterministic actions modifies a physical state:
    $$
    s' = \hat{a}\left(s\right)
    $$
    The action's effect on a belief state could be defined intuitively:
    $$
    m' = \hat{a}\left(m\right) := \bigcup_i{p_i \hat{a}\left(s_i\right)}
    $$
    Actions $\hat{A}$ with probabilistic outcomes $\hat{a}_j$ applied to a physical state $s$ result in a belief state:
    $$
    m' = \hat{A}\left(s\right) := \bigcup_j{p_j \hat{a}_j\left(s\right)}
    $$
    And a result of an action with probabilistic outcomes, applied to a belief state is also a belief state:
    $$
    m' = \hat{A}\left(m\right) = \bigcup_i{p_i \bigcup_j{p_j \hat{a}_j\left(s_i\right)}} =\bigcup_{i,j} {p_i p_j \hat{a}_j\left(s_i\right)}
    $$

    If we define an action by its postconditions (as in Eq. \ref{eq:actions-postc}), then setting postcondition values fully describes $\hat{a_j}$ functions.
    Corresponding Action node in BBT acts on a belief state as described above and returns a \textit{success} status.
    In the classical BT formulation, a node returns only the status to its parent.
    In BBTs, a node returns a full belief state and includes the returned status as variable $r_i$ to the belief state:
    \begin{equation}
        \hat{A}\left(m\right) = \bigcup_{i,j} {p_i p_j \{ f_j(s_i) | r_i := \msS \}}
        \label{eq:prob_action}
    \end{equation}

    \paragraph*{Conditions}
    Conditions are defined as a function ${c}(s)$ over a physical state which returns one of three statuses (\sS, \sF, and \sR). Condition node can be perceived as an action node that modifies only return status $r$ variable of each physical state. 
    $$
    {C}\left(m\right) = \bigcup_i {p_i \{ s_i | r_i := \hat{c}\left(s_i\right) \} }
    $$

    \subsection{Control nodes}
    Control node execution starts from a tick, applied to the first child. Working in belief space, we cannot return single status, as for different physical state there might be different execution results. 
    Hence, each node in the BBT returns a belief state instead of a single status (\sS, \sF~or \sR). For each physical state in returned belief state, we save returned status in variable $r_i$ ($i$ denotes iteration over physical states).
    In a BT, further execution of the control node's children depends on a returned status and therefore can vary among physical states.
    As it was noted in the previous work\cite{abtm}, Control nodes (Sequence, Fallback, and Skipper) rely on the same execution algorithm up to the \emph{return status} parameter. Hence, consider a Sequence node with all children being leaves.  Whenever the Sequence node receives a status of \sF~or \sR, it returns such state to its parent.
    Given that, in BBT for all physical states returned by the first child whether the $ r_i = \msS$, we should continue execution.
    Executing a Sequence node the BBT, we select all the physical states whether $r_i = \msS$, and pass them as an argument to the tick of second child. 
    All the remaining physical states with $r_i \neq \msS$, should not be passed but returned to the parent.
    We repeat the same procedure for all the children, passing a subset with $r_i = \msS$ to the next child and collecting all the other physical states for return.
    The whole returned belief state of the last child should be added up to the return of a Sequential control node.
    \newline
    Formally, for the recursive definition of the tick function we have to add two more arguments:
    \begin{itemize}

        \item \texttt{mem} - to pass a subset of Belief State
        \item \texttt{from} - to apply a tick from a \textit{i-th} child
    \end{itemize}
    Hence, we can obtain an elegant and compact definition of belief tick (Alg. \ref{alg:control-tick}). The tick function is now recursive not only with depth of the BT, but also in the left to right direction of the control node children. Therefore, it contains second argument \emph{from} that denotes the certain child of control node. If we the belief state argument is empty set, we return the empty set. In case we reached end of node's children list, we return the belief state argument. In other cases, we apply tick to child at the position \emph{from} in the node's children list. The result of tick is a belief state, that might contains different return statuses. Hence, we separate the physical states where the child node returned \sS status from the physical state where the child returned \sF or \sR. In the latter case Sequence node terminates the execution, so we add these states to returned belief state. \sS part of belief state we pass to the next child of the Sequence. Notice that an example of belief state flow chart is reported in Fig. \ref{fig:seq_tick}.
    \begin{algorithm}
        \caption{Tick function for Sequential}\label{alg:control-tick}
        \begin{algorithmic}[1]
            \Function{node.tick}{$mem$, $from = 0$}
            \If {$from \geq len(node.children)$ }
            \State \textbf{return} $mem$
            \EndIf
            \If {$mem = \mnt$}
            \State \textbf{return} $\mnt$
            \EndIf
            \State $mem = node.children[from].tick(mem)$
			\State $succeeded, failed = mem.split\_by(r_i = \msS)$
            \State \textbf{return} $failed + node.tick(succeeded, from + 1)  $ 
            \EndFunction
        \end{algorithmic}
    \end{algorithm}
    Calling a tick function on the root node with current belief state as an argument, the resulting belief state
    will fully describe all possible scenarios of tick propagation on all initial physical states.
    \subsection{Delayed action outcomes}
    In general, actions may imply that it takes some time to change the system's state while the tick execution in most works is a non-blocking procedure\cite{klockner2013}.
    If we directly apply actions in the way defined in Eq. \ref{eq:prob_action}, we would fork our belief state immediately before executing further BT nodes.
    That could result in a different behavior, compared to real BT execution.
    To avoid potential inconsistency, we \emph{delay} action outcomes by assigning a link to a delayed action to a specific state variable.
	If we never execute two or more actions in parallel, then the execution shall wait until action was finished. However, as we do not execute any action during this delay, we can assume for the simulation that an action finishes after one tick of the BT. So, we apply the action outcomes to belief state right before next tick.
    If the BT executes multiple actions in one tick e.g., by a Parallel node\cite{BTBook}, we could have multiple delayed actions simultaneously.
    Then, applying these actions on a belief state in a different order and with a different number of ticks in between could result in a different execution.
    Handling \emph{multiple} delayed actions and a Parallel node is out of the scope of this paper.
    \subsection {Self-simulation}
    \label{ss:self-ver}
    \begin{figure}[ht]
        \centering
        \includegraphics[width=0.8\columnwidth]{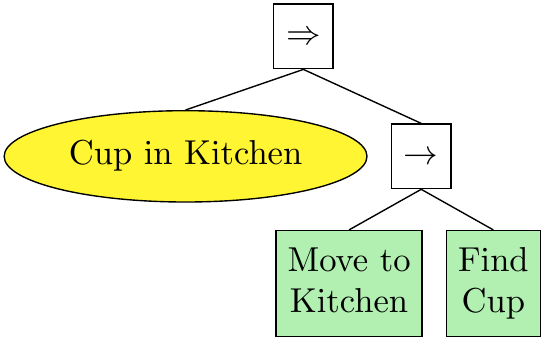}
         \caption{An extra sequence attached to the Skipper node imposes the desired behavior where the robot first moves to the kitchen and then finds the cup.}
         
         \label{fig:actions_for_skipper}
    \end{figure}
    We now define a \emph{self-simulation} procedure. Informally, it could be described as a forward belief statespace exploration procedure from an initial state applying a BBT policy.
        
    Giving the initial belief state and a BBT, we can simulate all the scenarios of execution.
    Note that the initial belief state could be simply a physical state with the probability of $1$ representing the current state of the robot.
    We define the self-simulation procedure limited to the cases when a robot could perform only one action at a time.
    For this reasons, we do not use Parallel node in this work, and always attach Action nodes to the Skipper through extra Sequence node to prevent tick passing to the further children (see the example on Fig. \ref{fig:actions_for_skipper}).
    So, we can apply a delayed action right after root tick finishes.
    Applying a tick to the root node and applying delayed action to the returned memory form a step of
    BBT \textit{self-simulation}.
    We can set a limit on number of ticks or the maximum number of states in a memory (the latter corresponds to the number of scenarios simulated).
    Notice that as we do not have any other source of memory changes, except for actions, and if for some physical state we do not have any delayed actions, this physical state shall be unchanged by further execution.
    Being executed on the same physical state, BT is guaranteed to reproduce the same tree traversal, and therefore, no actions shall be executed.
    Such physical states could be excluded from a belief state passed to the next root tick.
    Moreover, if all our actions are latched (i.e. once the action is finished, it shall not be executed again but return last status, \sS or \sF), the number of action nodes calls is limited to the number of actions in the tree.
    In this case, BT is guaranteed to finish execution in a finite number of ticks.
    \begin{algorithm}
        \caption{Self-simulation procedure}\label{alg:self-ver}
        \begin{algorithmic}[1]
            \Function{BT.simulate}{$mem$}
            \State $results = BeliefState([])$
            \While{\textbf{not} $mem.empty()$}
            \State $current = BT.root.tick(mem)$
            \State $ended, mem = current.split\_by( $
            \State ~$ s: s.has\_delayed\_actions())$
            \State $results = results + ended$
            \State $mem = mem.apply\_delayed\_actions()$
            \EndWhile\
    \State \textbf{return} $results$
            \EndFunction
        \end{algorithmic}
    \end{algorithm}

    \section{Automatic Synthesis of BTs}

    In this section, we present the third contribution of our paper.
    Having a BBT, we can trivially construct a corresponding BT because of each node type (Action, Condition, Control nodes) already has a BT definition.
    The execution of this BT will result in one of the physical states from the belief state.
    Therefore, we aim to construct first a BBT and then use corresponding BT for the runtime execution. For the inference we follow the definition in \cite{abtm}, which allows condition nodes to return running status.
    The planning pipeline was inherited from a previous work, that was aimed at BTs planning in a deterministic domain\cite{colledanchise2019towards}.
    Below we highlight important differences before formally describe the planning algorithm.
    Briefly, the main routine step is still to find a failed condition and resolve it by inserting nodes or permuting branches (Alg.\ref{alg:planning}).
    The planning is terminated when goal probability is achieved by means of self-simulation.

    \begin{algorithm}
        \caption{Planning routine}\label{alg:planning}
        \begin{algorithmic}[1]
            \Function{refine\_tree}{$initial\_state$, $bt$, $goal\_prob$}
            \State $bstate = BeliefMemory(state)$
            \State $prob = 0$
            \While{ $prob < goal\_prob$}
            \State $target = find\_failed\_condition(bt, bstate)$ \Comment{Section \ref{ss:find_cond}}
            \If {$threaten(target)$}
            \State ~$ resolve\_threat(bt, bstate, target)$
            \Else
            \State ~$ resolve\_by\_insert(bt, bstate, target)$ \Comment{Section \ref{ss:iter_insert}}
            \EndIf
            \State $bstate = bt.simulate(initial\_state)$
            \State $finished, bstate = bstate.split\_by(r_i = \msS)$
            \State $prob = finished.probability()$
            \EndWhile\
    \State \textbf{return} $results$
            \EndFunction
        \end{algorithmic}
    \end{algorithm}

    \subsection{Input of the planning problem}
    In the previous work\cite{colledanchise2019towards}, a set of goal conditions is an input to the planning problem.
    If we want to set an action as a planning goal, we can define a goal as set of this action preconditions.
    \par As we plan in a probabilistic domain, we should set a target success probability $p_{goal}$ and terminate our planning algorithm as soon as constructed BT is expected to succeed with probability  $p_{goal}$.
    In order to calculate it, as described in Section \ref{ss:self-ver}, we need to add an initial belief state for a planning problem. The current physical state of the robot could suite as an initial belief state for most applications.

    \subsection{Finding a condition to resolve}
    \label{ss:find_cond}
    \begin{figure}[ht]
        \centering
        \includegraphics[width=0.8\columnwidth]{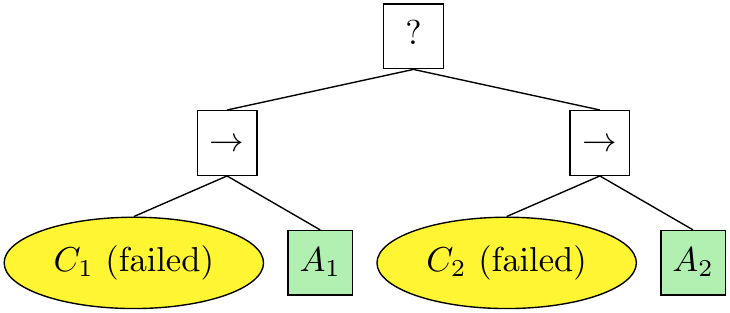}
        \caption{An example of the ambiguous choice of the condition to resolve.}
        \label{fig:cond_amb}
    \end{figure}
    At this step, we need to choose the condition that is either not satisfied and there is no action in the BT that satisfies it or the condition whose value is unknown and there is no sensing action that makes such value known. In general, there could be many such conditions in BT. Consider for example the Fallback node in Fig. \ref{fig:cond_amb}, in which the children are two actions with preconditions.
    If both actions have unsatisfied preconditions, we are unable to decide which action is preferred. In fact it might be that the one of the action's preconditions are easier to resolve: however this may be known only after further planning. Following previous on task planning with BT \cite{colledanchise2019towards}, we aim to find a \emph{deepest failed} condition.
    \par Assuming we ran a self-simulation procedure (Sec. \ref{ss:self-ver}), we end up in a belief state.
    Different scenarios of execution could infer different conditions to be ``the deepest failed". 
    Since we plan in belief space, for a BBT we have to choose the \textit{most probable deepest failed} condition over those that do not hold in physical states.
    As we construct a tree in a way to avoid cases mentioned above, by ``the deepest" we mean simply the deepest node in the tree.
    Hence, to find a condition to resolve $C_k$ we need to find the deepest condition with the highest cumulative probability over all physical states:
    $$
    k = argmax_i  \sum_j p_j I_{i,j}
    $$
    $I_{i,j} := 1$ if $C_i$ is the deepest failed in physical state $s_j$, $I_{i,j} := 0$ otherwise.

    \begin{figure*}[ht!]
        \begin{subfigure}[t]{0.38\textwidth}
                \centering
                \begin{subfigure}[t]{0.45\columnwidth}
                        \centering
                \includegraphics[width=0.99\columnwidth]{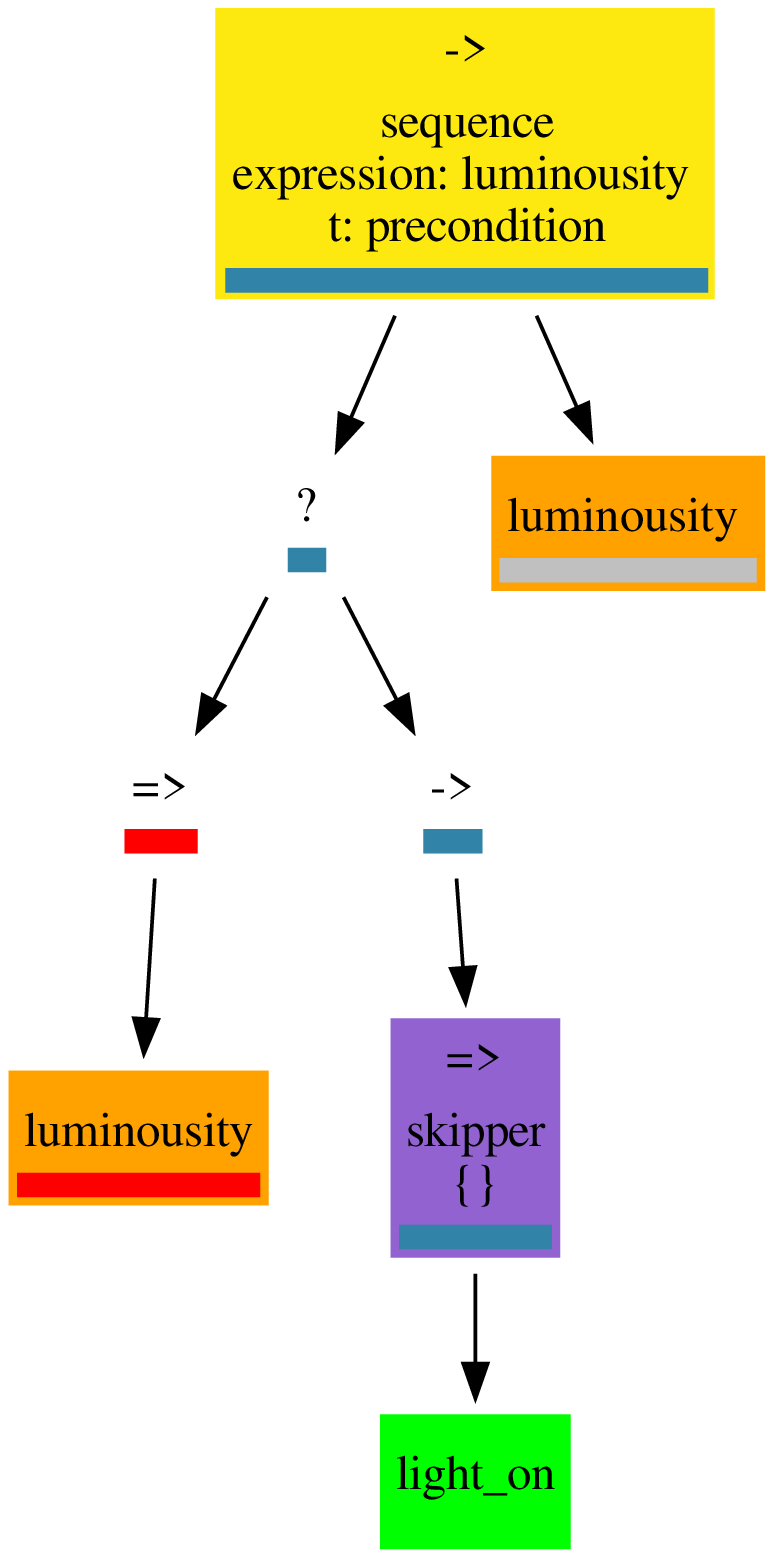}
                         \label{fig:rt1_bt}
                    \end{subfigure}%
                    ~
                    \begin{subfigure}[t]{0.34\columnwidth}
                \includegraphics[width=\columnwidth]{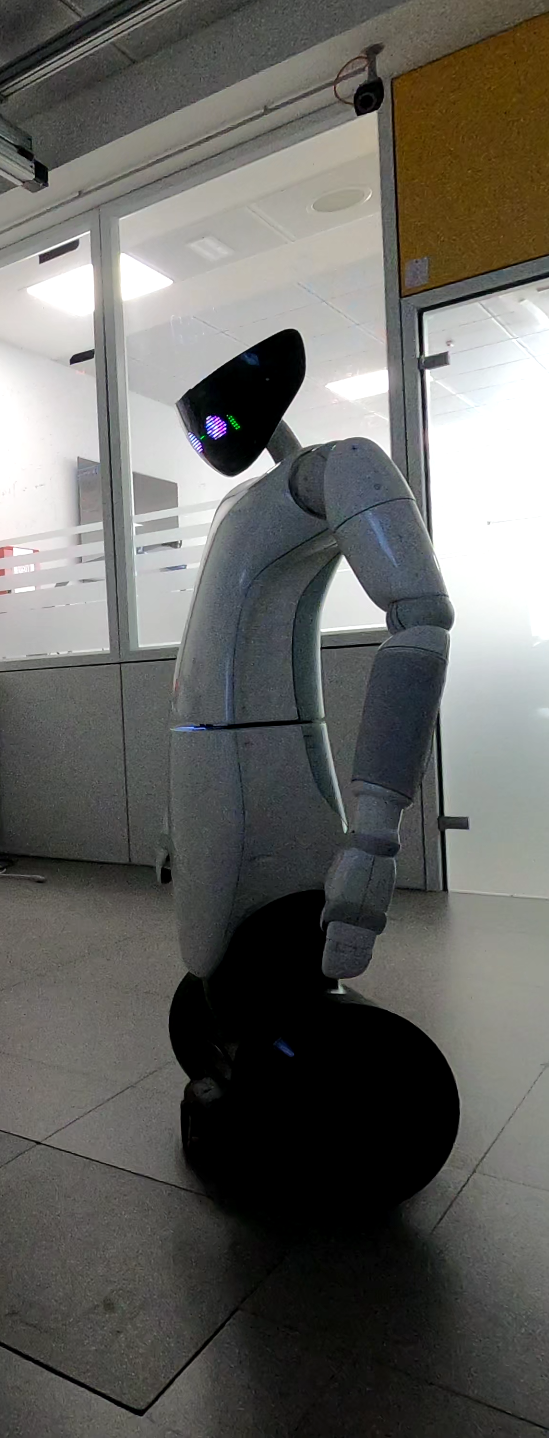}
                            \label{fig:rt1_R1}
                    \end{subfigure}
                    \vspace{0em}
            \end{subfigure}%
            ~
            \begin{subfigure}[t]{0.61\textwidth}
                \begin{subfigure}[t]{0.55\columnwidth}
                    \centering
            \includegraphics[width=0.99\columnwidth]{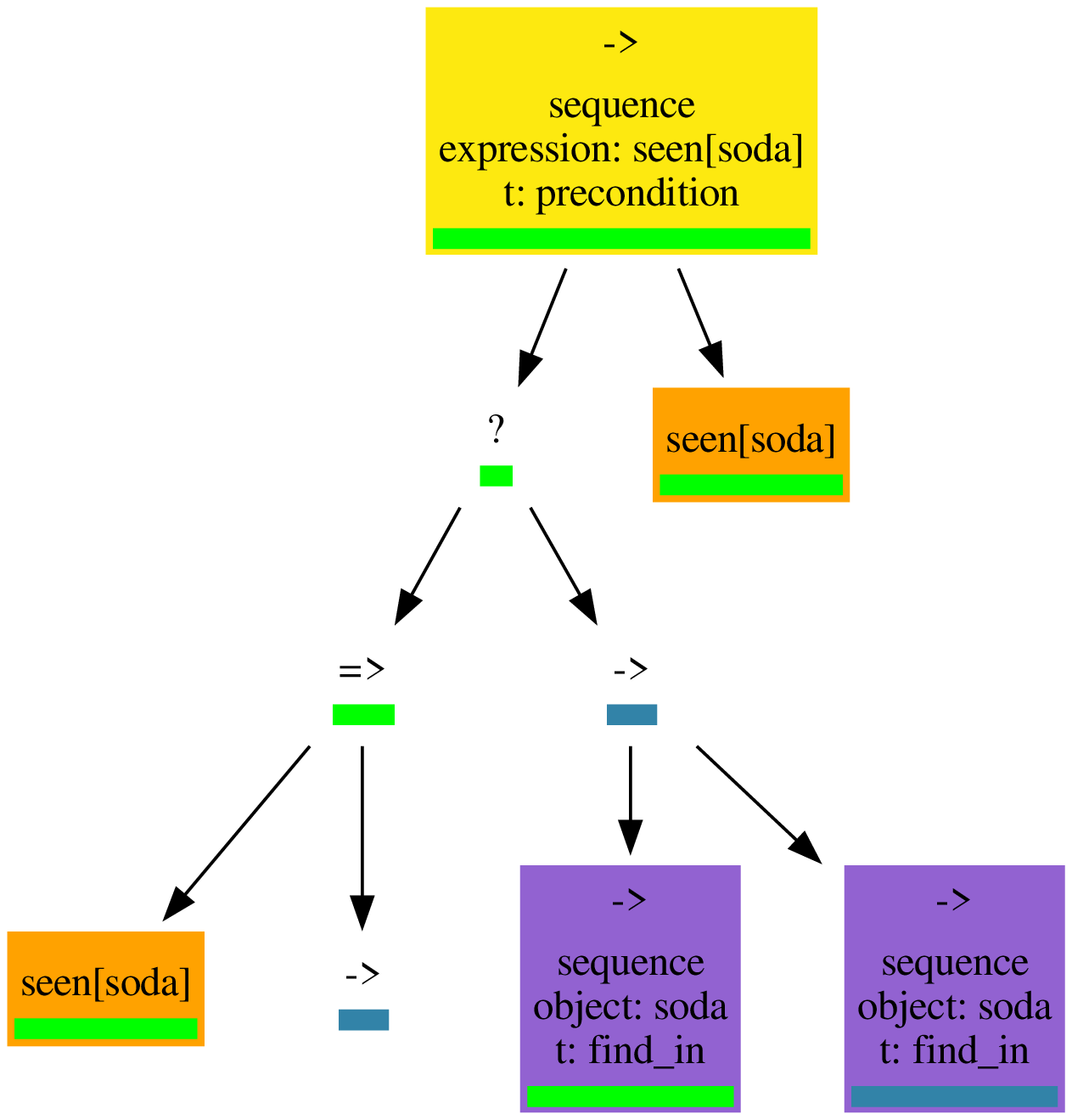}
                     \label{fig:rt2_bt}
                \end{subfigure}%
                ~
                \begin{subfigure}[t]{0.43\columnwidth}
            \includegraphics[width=\columnwidth]{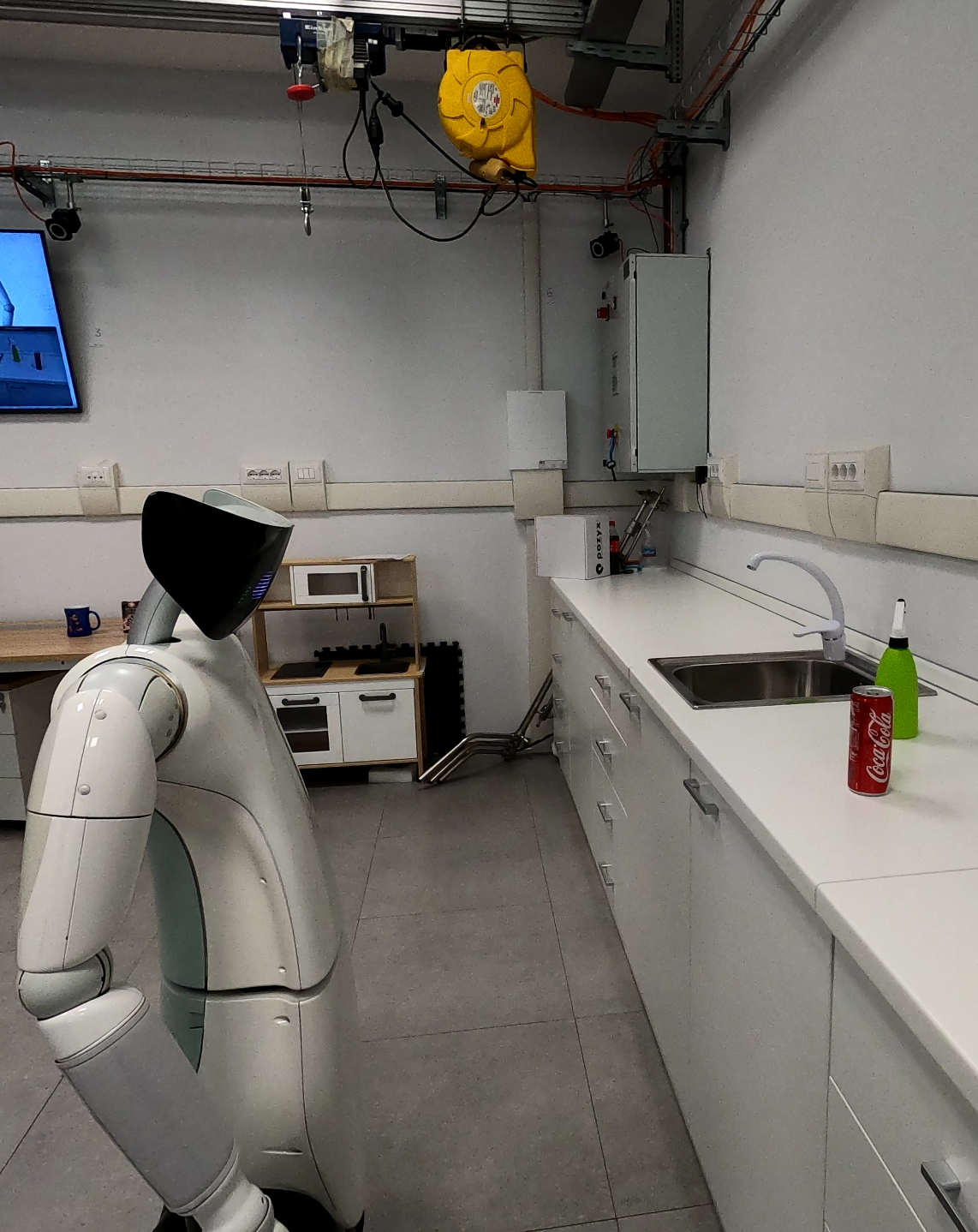}
                        \label{fig:rt2_R1}
                \end{subfigure}
                \vspace{0em}
            \end{subfigure}
            \vspace{0em}
            \caption{ 
            In this experiment R1 was asked to look for the object $soda$. We hid the object from the $table2$ during first execution of the $find$ node template.
            We show two R1 states, each represented by a screenshot of the part of the BT visualization that is executed on the robot and a photo of the R1 at this moment.
            On th first pair of pictures R1 detected poor luminousity condition ($luminousity\_ok = \msF$, condition node underscored with red on the screenshot) and commanded to turn on the light ($light\_on$).
            On the second pair of pictures we show an end of experiment, R1 is looking on the $soda$ can laying on the $table2$ place. Goal condition $seen(soda)$ is satisfied (returned \sS~status, underscored with green on the screenshot).
            Running nodes are represented with a blue bar, nodes with success and failure statuses are with a green and red bar respectively.
            }
            \label{figs:experiment}
    \end{figure*}

    \subsection{Iterative Behavior Tree Expansion}
    \label{ss:iter_insert}
    There are two possible reasons why target precondition could hold other than \sS~value.
    First, we check if there is a \emph{conflict}, a postcondition of previously inserted action coinciding with the target precondition.
    Following the literature~\cite{colledanchise2019towards}, we move tree branches to let this action be executed after the satisfaction of the target precondition.
    Then, if the precondition is still not satisfied or there was no conflict, we insert a \emph{latched} action to either a skipper node or a fallback node, depending on the value that the condition holds (correspondingly, \sR~or \sF). The latched node is an node, that. Moreover, one can prevent re\-executing a whole subtree using a latch node \cite{klockner2015behavior}.
    At least one of the action postconditions has to result in a target condition satisfaction.
    In case of several possible actions, we can choose the action to insert  by taking into account the following factors: 
    \begin{itemize}
    \item the probability of a successful postcondition
    \item the fact if all action's preconditions are satisfied in a current belief state 
    \item the history of previously executed actions
    \end{itemize} 
    Even though inserting \emph{latched} actions reduces the reactivity of BT, we find this approach more flexible.
    We can insert an arbitrary number of latched actions to achieve the goal success probability.
    Inserted actions could be different in case there is more than one action that has a target postcondition.
    In addition, we can insert not just actions, but \emph{node templates}, an arbitrary parameterized collection of nodes \cite{safronov2020node}.
    In the case when the result of a node template execution could be described simply by a set of postconditions,
    we can still assign postconditions to guide a planning algorithm.
    Any side effects shall be caught by self-simulation procedure and next steps of tree expansion.

	\begin{figure}[ht]
		\begin{subfigure}[t]{0.52\columnwidth}
        		\centering
			\includegraphics[width=0.99\columnwidth]{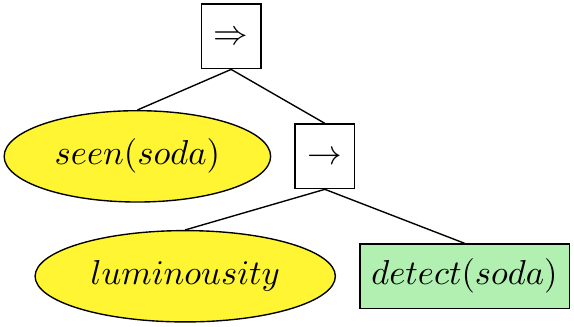}
         	\caption{We start from condition``seen(soda)''. In case it is unobservable, we insert a Skipper node with ``detect(soda)'' action, which checks if soda is detected right now.}
         \label{fig:example_find1}
    		\end{subfigure}%
    		~
        \begin{subfigure}[t]{0.43\columnwidth}
        		\centering
			\includegraphics[width=0.99\columnwidth]{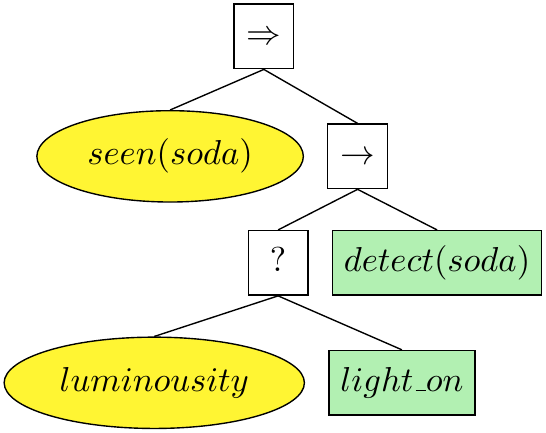}
         	\caption{It might happen that precondition ``luminousity'' of ``detect'' action could be unsatisfied. For this reason we insert action ``light\_on''. }
         \label{fig:example_find2}
    		\end{subfigure}%
    		\\
    		\begin{subfigure}[t]{0.97\columnwidth}
        		\centering
			\includegraphics[width=0.99\columnwidth]{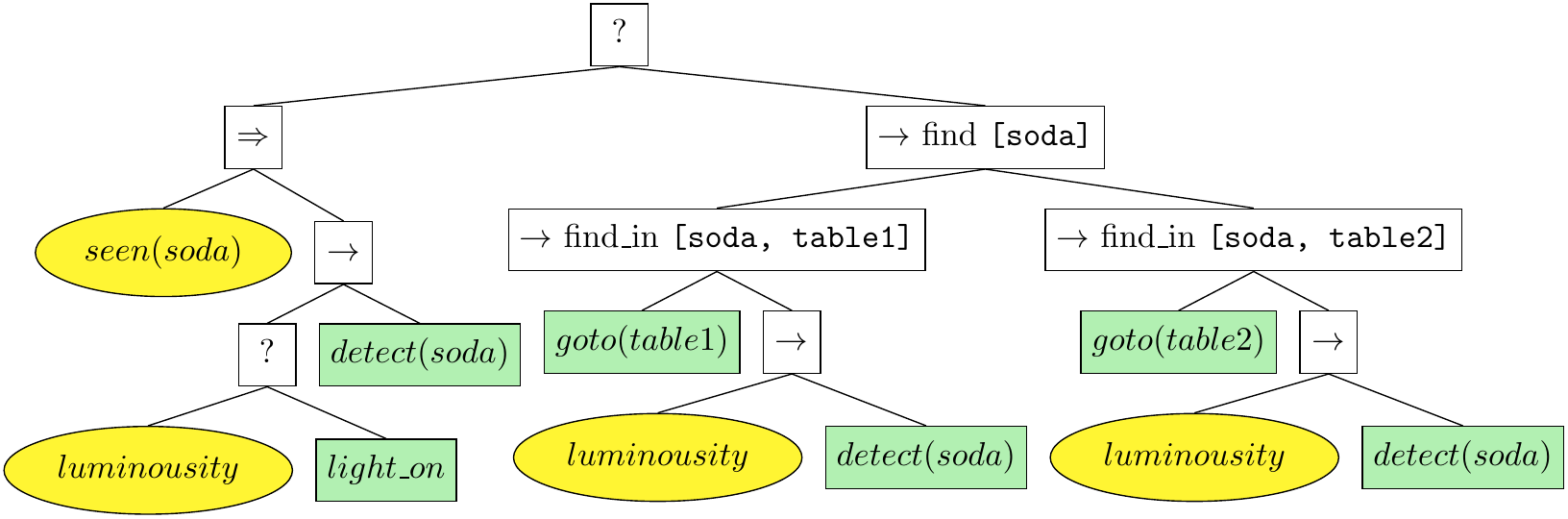}
         	\caption{In case we did not detect object ``soda'' on our current place, we search over possible locations. It is implemented by template ``find'' which consists of templates ``find\_in''. Note, that we do not need to resolve condition``luminousity'' anymore, as it must be already resolved by action ``light\_on'' with probability 1 (check Table \ref{tbl:actions}.}
         \label{fig:example_find3}
    		\end{subfigure}%
    		\\
        \begin{subfigure}[t]{0.97\columnwidth}
        \centering
        \includegraphics[width=0.99\textwidth]{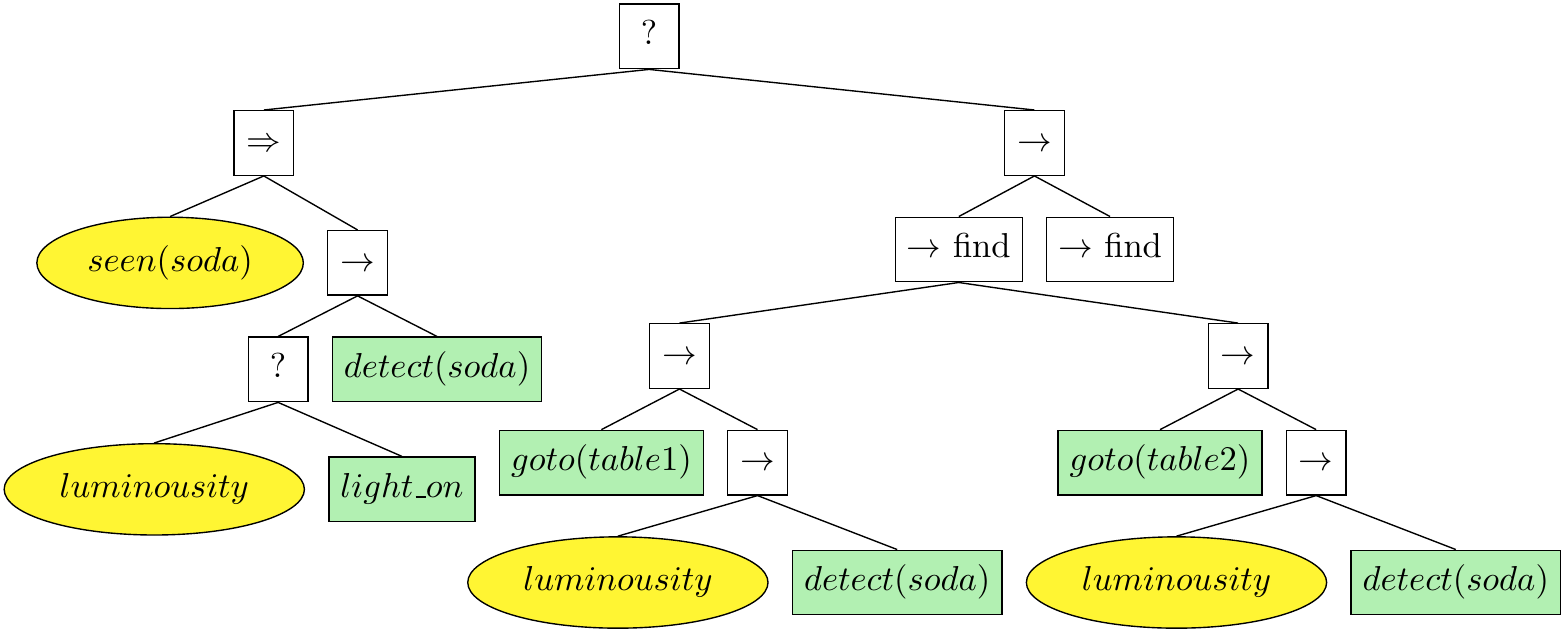}
        \caption{Final BT solution includes another attempt to find ``soda'' object. The second subtree ``find'' is not expanded on the picture to keep it compact.}
        \label{fig:example_find}
        \end{subfigure}
        \caption{An example of iterative BT expansion step by step for the ``look for soda'' scenario on Fig. \ref{figs:experiment}}
        \vspace{-1em}
	\end{figure}
    \section{Experimental Validation}
    In this section, we report the experimental validation. Framework implementation, specification of action and condition sets, and the example below are available online\footnote{https://github.com/safoex/pybt}.
    To validate our planning approach, we constructed a set of actions and conditions, that relates to the domestic application scenario. We used an off-the-shelf object detection pipeline to implement the condition $seen(object)$ that checks if an object is in the robot's  camera field of view~\cite{maiettini2018,maiettini2019a}.
    We executed the generated BTs on the R1 robot~\cite{parmiggiani17} and achieved correct execution.
    For simplicity, in the paper we describe a subset of this domain and a simple goal.
    The domain is described by sets of actions (including node templates) (see Table \ref{tbl:actions}),
    conditions (see Table \ref{tbl:conditions}), and their parameter spaces (see Table \ref{tbl:parameters}).

    \begin{table}[h!]
        \centering
    \begin{tabular}{|l | c | c | c|}
    \hline
    Action           & Params	& Preconditions & Postconditions 	        \\ \hline
    \scriptsize goto 			& \scriptsize place         &  	\nt       & \scriptsize  $0.95, at(place) = \msS$ \\
         			&               &  	            &  \scriptsize $0.05, no~ changes$       \\ \hline
    \scriptsize detect 	        & \scriptsize object 		& \scriptsize $luminousity\_ok = \msS$	& $0.5, seen(object) = \msS$\\
         	        &        		& \scriptsize $seen(object) = \msR$  & $0.5, seen(object) = \msF$\\ \hline
    \scriptsize light\_on       &               &   \nt       & \scriptsize $1, luminousity\_ok = \msS$   \\ \hline
    \scriptsize find            & \scriptsize object   & \scriptsize $seen(object) = \msF$ & $0.8, seen(object) = \msS$ \\
                    &               &               & \scriptsize $0.2, seen(object) = \msF$\\ \hline
    \end{tabular}
            \caption{Actions}
        \label{tbl:actions}
    \end{table}

    \begin{table}[h!]
        \centering
        \begin{subfigure}[t]{0.6\columnwidth}
            \centering
                \begin{tabular}{|l | c | c | c|}
                \hline
                Condition        & Parameters & Values \\ \hline
                at 		    	& place      & \sS, \sF   \\ \hline
                seen            & object     & \sS, \sF, \sR  \\ \hline
                luminousity\_ok  & \nt        & \sS, \sF \\ \hline
                \end{tabular}
                \caption{\emph{Values} contains \sR for conditions which might be unobservable.}
                \label{tbl:conditions}
        \end{subfigure}%
        ~
        \begin{subfigure}[t]{0.39\columnwidth}

            \centering
            \begin{tabular}{|l | c | c | c|}
            \hline
                            & Instances	\\ \hline
            place 		    & table1, table2    \\ \hline
            object          & soda, sprayer     \\ \hline
            \end{tabular}
            \caption{Parameters}
            \label{tbl:parameters}
        \end{subfigure}
        \vspace{0em}
        \caption{
        Conditions (a) and parameters (b), used in our validation scenario.  }
        \label{tbls:pars_and_cond}
    \end{table}
    
    \begin{table}[h!]
    		\centering
    		\begin{tabular}{|c | c |}
                \hline
                Nodes      & Return statuses \\ \hline
                \colorbox{orange}{condition}  	& \colorbox{grey}{never started yet}       \\ \hline
                \colorbox{green}{action}    & \colorbox{running}{running}       \\ \hline
                \colorbox{tempexp}{template (expanded)}  &  \colorbox{red}{failed}       \\ \hline
                \colorbox{tempnot}{template (not expanded)}  &  \colorbox{green}{success}       \\ \hline
        \end{tabular}
        \caption{Color legend for BT on Fig. \ref{figs:experiment}}
    \end{table}

    The robot has to find a soda can with a probability of $0.9$. The goal of this task is described by the condition $seen(soda)$ and goal probability was set to $0.9$.
    The initial state of the robot was $seen(soda) = \msR, at(table1) = \msF, at(table2) = \msF, luminousity\_ok = \msF$ (omitting unused conditions).
    Let us follow the planner step by step. As the $seen(soda)$ was unobserved, the planner inserted the \emph{perception} action $detect$ (see Fig. \ref{fig:example_find1}).
    In order to make this observation, the condition $luminousity\_ok$ should hold $\msS$.
    Therefore, $light\_on$ action was inserted (see Fig. \ref{fig:example_find2}). After this step, the success probability is $0.5$.
    As the target probability was set higher, and all failed states contained $seen(soda) = \msF$ condition, we inserted a \emph{find} sub-BT (see Fig. \ref{fig:example_find3}).
    After this, success probability achieved $0.875$.
    Note, that we did not need to insert extra $light\_on$ actions because in all paths of execution,
    the precondition $luminousity\_ok = \msS$ was already successfully satisfied.
    After  the second $find$ template was inserted (not expanded on the Fig.\ref{fig:example_find}),
    the success probability reached $\simeq 0.97$, and planning was terminated.
    Note, that setting the target probability closer to $1$ would force the planner to insert more and more $find$ templates, pushing robot to make more and more attempts to find an object. 
    Such behavior corresponds to one \emph{not latched} node template $find$, which could be a result of planning in a not probabilistic domain.

    \section{Conclusion}
    In this paper, we proposed a planning approach to automatically create BTs that takes into account state uncertainty.  We extended the formulation of condition nodes, allowing them to represent the situation in which the value of a condition is unknown. This allowed us to handle both current and future state uncertainty. Our approach combines modularity and reactivity of BTs with automated planning.
    Planning in a probabilistic domain allowed us to control the target success probability.
   We demonstrated successful real robot execution of BT synthesized by our algorithm.
    
    \bibliography{root}

    \bibliographystyle{ieeetr}
\end{document}